\documentclass[11pt,a4paper,dvipsnames]{article}
\usepackage[hyperref]{emnlp2020}
\usepackage{times}
\usepackage{latexsym}
\usepackage{graphicx}
\usepackage{booktabs}
\usepackage{csquotes}
\usepackage{amsmath,amsfonts,amssymb}
\usepackage[euler]{textgreek}
\usepackage{multirow}
\usepackage{enumitem}
\usepackage{algorithm}
\usepackage{algorithmic}

\usepackage{microtype}

\aclfinalcopy % Uncomment this line for the final submission
\newcommand{\papertitle}{Blank Language Models}
\title{\papertitle}

\author{
Tianxiao Shen\footnotemark[1] \quad Victor Quach\footnotemark[1] \quad Regina Barzilay \quad Tommi Jaakkola \\
MIT CSAIL \\
\texttt{\{tianxiao, quach, regina, tommi\}@csail.mit.edu}
}

\date{}

\begin{document}
\maketitle
\renewcommand*{\thefootnote}{\fnsymbol{footnote}}
\footnotetext[1]{Equal contribution}
\renewcommand*{\thefootnote}{\arabic{footnote}}
\begin{abstract}
We propose Blank Language Model (BLM), a model that generates sequences by dynamically creating and filling in blanks.
The blanks control which part of the sequence to expand, making BLM ideal for
a variety of text editing and rewriting tasks.
The model can start from a single blank or partially completed text with blanks at specified locations.
It iteratively determines which word to place in a blank and whether to insert new blanks, and stops generating when no blanks are left to fill.
BLM can be efficiently trained using a lower bound of the marginal data likelihood.
On the task of filling missing text snippets, BLM significantly outperforms all other baselines
in terms of both accuracy and fluency.
Experiments on style transfer and damaged ancient text restoration demonstrate the potential of this framework for a wide range of applications.\footnote{Our code is available at \url{https://github.com/Varal7/blank_language_model}}
\end{abstract}
\section{Introduction}

Neural language models have shown impressive performance across many applications such as machine translation and summarization where the text is generated from scratch \citep{bahdanau2014neural,rush2015neural}.
However, a broader set of text generation tasks --- including text editing, information fusion, and ancient text restoration --- requires the model to start with partially specified text and generate the missing fragments.
In the general setup, the input document may have any number of missing spans, and each span may have an unknown number of missing tokens.
To perform this text infilling task \cite{zhu2019textinfilling}, a model should:
(1) provide fine-grained control over the generation location,
(2) accommodate a variable number of missing tokens,
and (3) respect both the preceding and following context.

% variable length 
% respect context (consistency)
% self-consistent 
% valid

%try to adjust left-to-right language models to perform text infilling, but encounter inherent difficulties in meeting all three desiderata.
%To address \emph{consistency}, previous work introduced intricate inference algorithms, such as dynamic programming or gradient search, to find the filling content that has a high likelihood within the surrounding context  \citep{sun2017bidirectional,liu2019-tigs,Zaidi2020Decoding}.
%These methods are limited to the fixed-length setting.
%Moreover, they have to make simplifications like Markov assumptions and require high time complexity for decoding.

%A readily available option for this task is to utilize existing left-to-right language models.
Existing approaches focus on adapting left-to-right language models for text infilling. 
Intricate inference algorithms leveraging dynamic programming or gradient search are proposed to find the filling content that has a high likelihood within the surrounding context \citep{sun2017bidirectional,liu2019-tigs,Zaidi2020Decoding}. 
These methods make simplified Markov assumptions, require high decoding time complexity,
and cannot adapt to variable infilling length.
Alternatively, \citet{donahue2020enabling} predict the concatenation of the infilling content, but do not guarantee that the output will match the number of missing spans in the input.

\begin{figure}[t]
\fontsize{10}{12}\selectfont
%\begin{displayquote}
\hspace*{10pt} They also have \underline{\qquad} which \underline{\qquad} .\\
\hspace*{10pt} They also have \underline{\textit{ice cream}} which \underline{\textit{is really good}} .
%\end{displayquote}
%\vspace{-5pt}
\caption{BLM fills in blanks of arbitrary length.}\label{fig:example_infill}
%\vspace{-5pt}
\end{figure}

\begin{figure*}[t]
\centering
\fontsize{10}{12}\selectfont
\begin{tabular}{cl clcc}
 \toprule
  & \multicolumn{1}{c}{Canvas $c$}  &  \multicolumn{4}{c}{Action $a$}\\
  Step $t$      &    \multicolumn{1}{l}{}    & Location $b$  & Word $w$ & (Left blank $l$, & Right blank $r$) \\
 \midrule
 0.& \underline{\quad \#1 \quad} & \#1 & is &  Yes & Yes \\
 1.& \underline{\quad\#1\quad} is \underline{\quad\#2\quad} & \#1 & customer & No & Yes \\
 2.& customer \underline{\quad\#1\quad} is \underline{\quad\#2\quad} & \#2 & awesome & No & No \\
 3.& customer \underline{\quad\#1\quad} is awesome & \#1 & service & No & No \\
 4.& customer service is awesome & \multicolumn{4}{c}{-End-}\\
 \bottomrule
\end{tabular}
\iffalse
\begin{tabular}{cl clcc}
 \toprule
  & \multicolumn{1}{c}{Canvas $c$}  &  \multicolumn{4}{c}{Action $a$}\\
  Step $t$      &    \multicolumn{1}{l}{}    & Location $b$  & Word $w$ & (Left blank $l$, & Right blank $r$) \\
 \midrule
 0.& \underline{\quad \#1 \quad} & \#1 & have &  Yes & Yes \\
 1.& \underline{\quad\#1\quad} have \underline{\quad\#2\quad} & \#1 & They & No & Yes \\
 2.& They \underline{\quad\#1\quad} have \underline{\quad\#2\quad} & \#2 & . & Yes & No \\
 3.& They \underline{\quad\#1\quad} have \underline{\quad\#2\quad} . & \#2 & which & Yes & Yes \\
 4.& They \underline{\quad\#1\quad} have \underline{\quad\#2\quad} which \underline{\quad\#3\quad} . & \#1 & also & No & No \\
 5.& They also have \underline{\quad\#1\quad} which \underline{\quad\#2\quad} . & \#2 & really & Yes & Yes \\
 6.& They also have \underline{\quad\#1\quad} which \underline{\quad\#2\quad} really \underline{\quad\#3\quad} . & \#1 & ice & No & Yes \\
 7.& They also have ice \underline{\quad\#1\quad} which \underline{\quad\#2\quad} really \underline{\quad\#3\quad} . & \#2 & is & No & No \\
 8.& They also have ice \underline{\quad\#1\quad} which is really \underline{\quad\#2\quad} . & \#1 & cream & No & No \\
 9.& They also have ice cream which is really \underline{\quad\#1\quad} . & \#1 & good & No & No \\
 10.& They also have ice cream which is really good . & \multicolumn{4}{c}{-End-} \\
 \bottomrule
\end{tabular}
\fi
\caption{An example trajectory that generates the sentence ``customer service is awesome''. Each action is a tuple $(b, w, l, r)$, indicating the blank location $b$ selected for expansion, the word $w$ to fill in, whether to create a left blank $l$, and whether to create a right blank $r$.}\label{fig:example_trajectory}
\end{figure*}

In this work, we introduce the Blank Language Model (BLM), %, which supports flexible text generation, adapts to targets of variable length, and guarantees valid text infilling with straightforward decoding strategies.
which uses a special ``\underline{\quad}'' symbol to control where tokens can be placed.
The generation of BLM follows the grammar of replacing a blank with a word and possibly adjoining blanks.
By jointly modeling context and missing content, BLM supports the control of generation location and produces consistent infilling of variable length.

Our model can start from a single blank or partial text with blanks in specified locations. 
It maps the entire input into a sequence of vector representations, and further processes the representations in blank positions to determine the generation action.
Generation actions are performed iteratively until there are no blanks.
%Our BLM is based on a Transformer encoder that maps the input %text containing blanks
%into a sequence of vector representations.
%The representations at blank locations are further processed to %select a blank, a word to fill in it, and whether to generate adjoining blanks.
%determine the generation action.
%RB You are using the term trajectory but define it later 
Since multiple trajectories of BLM actions can produce the same final text, we train the model by maximizing a lower bound of the log-likelihood marginalized over trajectories.
%To make training more efficient, and to introduce an inductive bias towards order independence, we maximize instead a lower bound on the marginal likelihood.
At test time, we can use simple greedy decoding or beam search to fill in the blanks in the input text.

BLM shows superior performance in text infilling~\citep{zhu2019textinfilling}, ancient text restoration~\citep{assael-etal-2019-restoring} and style transfer~\citep{shen2017style}, demonstrating its flexibility to generate text in diverse conditions.
%RB This result sounds very small, esp. in light of the previous %sentence that talks about superior performance
Our model achieves 92.5\% accuracy and BLEU score of 23.1 on the Amazon dataset for sentiment transfer.
On the task of restoring ancient text that lost half of the characters, we reduce the error rate by 3.3 points compared to previous methods.
%In the restoration of ancient text that lost half of the characters, we reduce the error rate by 3.3 points compared to previous methods.

% on language modeling, and obtain perplexity comparable to left-to-right language models on Penn Treebank and WikiText datasets.

\section{Related Work}

% Preliminary work on text infilling has been restricted to the case where the length of the blank to complete is fixed and given \citep{joshi2020spanbert, liu2019-tigs}.
% Other dedicated systems for text infilling learn to predict the content of the blanks in a sequence-to-sequence fashion \citep{donahue2020enabling,fedus2018maskgan} but fail to guarantee 100\% valid predictions.

%Alternatives to conventional left-to-right generation have previously been explored from multiple approaches.
%Multiple prior approaches explored alternatives to conventional left-to-right generation.
%Part of these efforts was focused on finding an optimal generation order,
%including syntax-based approaches and methods for learning  adaptive generation order~\citep{emami2005neural,zhang2015top,dyer2016recurrent,ford2018importance,zhou2019synchronous,welleck2019non,gu2019insertion}. 

Recent work has explored various sequence models for non-autoregressive machine translation \citep{gu2017non}. %,lee2018deterministic,stern2019insertion,gu2019levenshtein}.
The Insertion Transformer supports dynamic canvas with word insertion \cite{stern2019insertion}, but does not allow users to specify where to insert.
The model is unaware of which parts of the canvas are contiguous text spans that should remain intact, and which (potentially scattered) parts need to be filled in.
Directly forcing the Insertion Transformer to perform text infilling can therefore lead to suboptimal solutions.
The Levenshtein Transformer combines insertion and deletion through complex policy learning
%and cannot efficiently estimate the probability of generating a sentence
\citep{gu2019levenshtein}.
Its insertion mechanism is a two-stage process in which placeholders are first predicted and then filled-in in a masked language model (MLM) manner.
In text infilling where the blanks/placeholders are given, it reduces to an MLM.

%Additional insertion control is provided by the masked language model where each mask corresponds to a single word~\citep{fedus2018maskgan}.
MLMs are commonly used in representation learning~\citep{devlin2018bert, joshi2020spanbert}.
To use them in rewriting tasks, one needs to specify the insertion length in advance
and heuristically determine the generation order among the masks~\citep{fedus2018maskgan,wang2019bert,ghazvininejad2019constant}.
Similarly, XL-Net requires absolute positional embedding and thus does not support unknown-length text infilling~\citep{yang2019xlnet,shih2019xl}.
%A blank in our model can correspond to any number of words, thereby avoiding the problem of predicting length.
BLM provides a natural formulation for generative modeling that can dynamically accommodate insertions of various length.

Another line of work focuses on finding an optimal language generation order, such as syntax-based generation \citep{dyer2016recurrent} and learning adaptive generation order \citep{gu2019insertion}. These approaches are tailored to generation from scratch in a specific order. Our model instead is attuned for text rewriting, where the missing parts can be located anywhere in the input text, and the algorithm must flexibly complete them.

\section{Blank Language Models}

\begin{figure*}[t]
\centering
\includegraphics[width=0.98\textwidth]{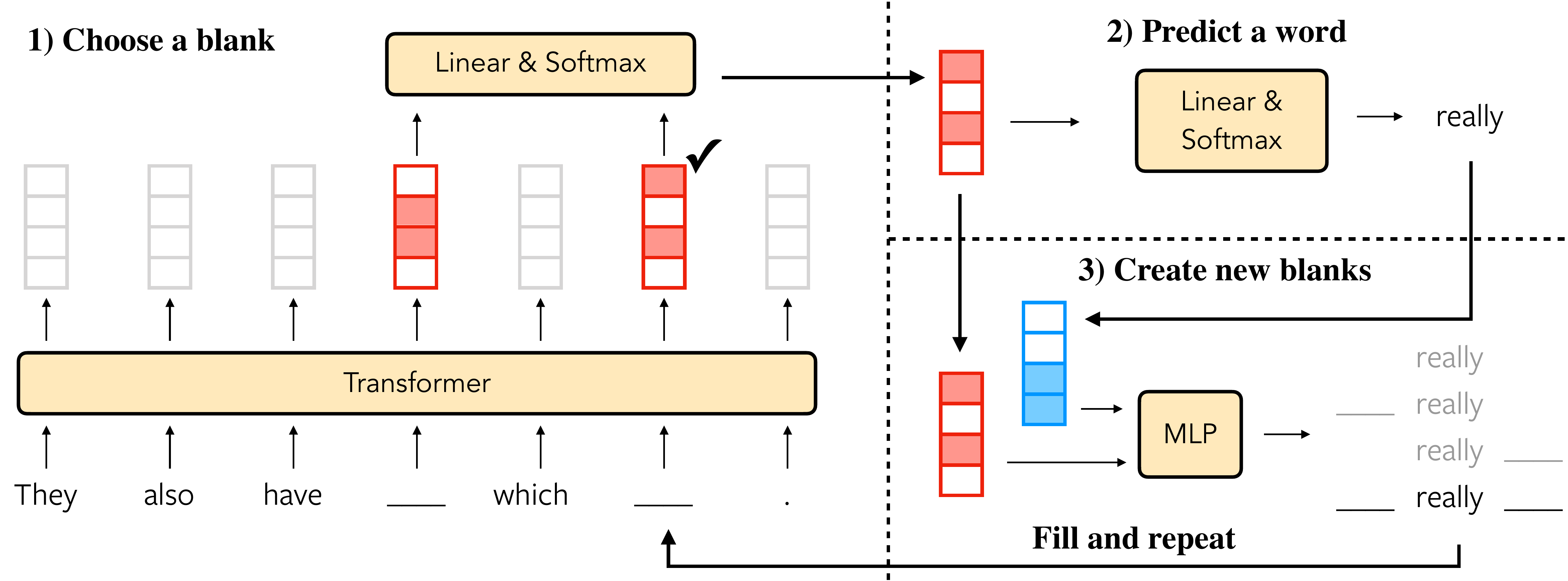}
\caption{ 
Architecture of the BLM.  In the first stage, an index is chosen among all current blank positions. For that location, a word is selected in the second stage. In the final stage, the blank representation is concatenated with the chosen word's embedding and fed into an MLP to determine the creation of the following blanks.
}\label{fig:model}
\end{figure*}

A blank language model (BLM) generates sequences by creating and filling in blanks.
Generation starts with a single blank and ends when there is no blank.
In each step, the model selects a blank ``\underline{\quad}'', predicts a word $w$, and fills the blank with ``$w$'', ``\underline{\quad} $w$'', ``$w$ \underline{\quad}'', or ``\underline{\quad} $w$ \underline{\quad}''.
This way, a blank can be expanded to any number of words.%, different from masks in MLM that correspond to exactly one word.
%Fig.~\ref{fig:example_action} shows an example action in one step, and
%Fig.~\ref{fig:example_trajectory} illustrates the entire generating process.

We define a \emph{canvas} as a sequence of words interspersed with special ``\underline{\quad}'' tokens. The subsequent action is conditioned on this intermediate stage of generation.
%Different from the Insertion Transformer that can insert words anywhere in between existing tokens, the BLM will only place words on the specified blanks.
Suppose the current canvas is $c=(c_1,\cdots,c_n)$
with blanks located at indices $b_1,\cdots,b_k$ (i.e.  $c_{b_i}=$ ``\underline{\quad}'', for $i=1,\ldots,k$). BLM maps this canvas to a distribution over actions specifying how the canvas is to be revised: 
\begin{align}
    p(b, w, l, r | c;\theta)=\text{BLM}(c)
\end{align}
where $b\in \{b_1,\cdots,b_k\}$ is a blank location; $w$ is a word in the vocabulary $V$; $l,r \in\{0, 1\}$ denote whether or not to create a blank to the left and right of $w$; and $\theta$ are the model parameters. The \emph{action}, defined as the tuple $(b,w,l,r)$ uniquely specifies the next state of canvas (see Fig.~\ref{fig:example_trajectory} for illustration).

Alternatively, we can view the actions in BLM as production rules in a grammar. Each blank represents a nonterminal symbol or the start symbol, and the terminal symbols come from the vocabulary $V$. The production rules are restricted to be of the form ``\underline{\quad}'' $\rightarrow$ ``\underline{\quad}?$w$\underline{\quad}?'' for $w\in V$, where ``?'' indicates that the preceding symbol is optional. In contrast to context-free grammars, the probability distribution over production rules in BLM is conditioned on the entire canvas generated so far.

\paragraph{Model Architecture}

We encode the canvas $c$ into a sequence of representations $(z_1,\cdots,z_n)$, and take representations $Z=(z_{b_1},\cdots,z_{b_k})$ where the blanks are located. Let $d$ denote the dimension of $z$'s.
We factorize the joint distribution $p(b, w, l, r | c;\theta)$ into three parts (shown in Fig.~\ref{fig:model}):
\begin{enumerate}[topsep=3pt]
\setlength\itemsep{0pt}
    \item Choose a blank:
    \begin{align}
    p(b_i|c; \theta)=\text{Softmax}(u^TZ)
    \end{align}
    where $u\in \mathbb R^d$ is a parameter vector to project $z$'s into one-dimensional logits.

    \item Predict a word for the selected blank:
    \begin{align}
    p(w|c, b_i; \theta)=\text{Softmax}(Wz_{b_i})
    \end{align}
    where $W\in \mathbb R^{|V| \times d}$ is a parameter matrix to project $z_{b_i}$ into the vocabulary.

    \item Decide whether or not to create blanks to the left and right of the predicted word:
    \begin{align}
    p(l, r|c, b_i, w; \theta)=\text{MLP}(z_{b_i}, v_w)
    \end{align}
    where $v_w$ is the word vector of $w$, and MLP is a multilayer perceptron with $4$ output classes: Left.Yes/No $\times$ Right.Yes/No.
\end{enumerate}

\paragraph{Likelihood}
Now let us consider the probability $p(x;\theta)$ of generating a sentence/paragraph $x=(x_1,\cdots,x_n)$ under the BLM.
We call the generating process from an initial blank to complete text a \emph{trajectory}.
The same final text $x$ can be realized by multiple trajectories. However, if we specify the order in which the words in $x$ are generated, the trajectory will be uniquely determined.
%This follows from the fact that BLM never induces a canvas with two (or more) consecutive blanks. 
%RB The point about generation order and trajectories seems quite clear. I am not sure if this long example is needed. But maybe I missed your point here.
Consider the example trajectory of a $4$-word sentence in Fig.~\ref{fig:example_trajectory}. Given the order $(3, 1, 4, 2)$, at step $0$ when we generate $x_3$, both left and right blanks are created for future generations of $x_1$ and $x_2, x_4$.
In step $1$ of generating $x_1$, only a right blank is created for the future $x_2$.
Subsequent steps can be deduced by analogy. The correspondence between trajectories and generation orders allows us to write the marginal likelihood as:
\begin{align}
    p(x;\theta) &= \sum_{\sigma\in S_n} p(x,\sigma;\theta) \nonumber \\
    &= \sum_{\sigma\in S_n} \prod_{t=0}^{n-1} p(a^{x,\sigma}_t|c^{x,\sigma}_t ;\theta)
    \label{eq:likelihood}
\end{align}
where $S_n$ is the set of all $n$-permutations; $a^{x,\sigma}_t,c^{x,\sigma}_t$ denote the action and canvas at step $t$ under sentence $x$ and order $\sigma$, respectively (cf. Fig.~\ref{fig:example_trajectory}).

\paragraph{Training}
Different losses have been proposed to train generalized sequence models.
%RB I am not sure I understood the part about 15%. I reread it multiple times -- still can't get it
For instance, BERT and XL-Net mask and predict 15\% of tokens conditioned on the rest. This strategy is more suitable for representation learning rather than generation.
Insertion Transformer masks different numbers of tokens and weights them with uniform loss or binary tree loss~\citep{stern2019insertion,chan2019kermit}. It aims to perform fast inference through parallel decoding.
Here, we present a training objective from the language modeling perspective by estimating the log likelihood of generating $x$.
%It also enables us to evaluate the perplexity of BLM and compare with traditional language models.

\begin{algorithm}[t]
\caption{BLM training\footnotemark}
\label{alg:train}
\fontsize{9.5}{11.4}\selectfont
\begin{algorithmic}[1]
\STATE Initialize model parameters $\theta$
\REPEAT
 	\STATE Sample a training example $x = (x_1,\cdots,x_n)$\\
    \STATE Sample $t$ from $0$ to $n-1$\\
    \STATE Sample an $n$-permutation $\sigma$\\
 	\STATE Construct canvas $c$ that keeps tokens $x_{\sigma_j} (j = 1, \cdots, t)$ and collapses remaining tokens as blanks\\
 	\STATE Get $n-t$ target actions $a_{j-t}$ for filling $x_{\sigma_j}$ $(j = t+1, \cdots, n)$ into canvas $c$\\
    \STATE Compute $\text{loss}(\{a_1, \cdots, a_{n-t}\}, \text{model}.\text{forward}(c))$ from Eq.~(\ref{eq:loss})\\
 	\STATE Update $\theta$ by gradient descent
\UNTIL{Convergence}
\end{algorithmic}
\end{algorithm}
\footnotetext{We implement a batch version of the algorithm.}

Directly computing the marginal likelihood over $n!$ orders is intractable.
We apply Jensen's inequality to lower bound the log likelihood:
\begin{align}
    \log p(x;\theta) = \log \sum_{\sigma\in S_n} \prod_{t=0}^{n-1} p(a^{x,\sigma}_t|c^{x,\sigma}_t ;\theta) \nonumber \\
    %\ge \log(n!) + \frac{1}{n!} \sum_{\sigma\in S_n} \log\prod_{t=0}^{n-1} p(a^{x,\sigma}_t|c^{x,\sigma}_t ;\theta) \nonumber \\
    \ge \log(n!) + \frac{1}{n!} \sum_{\sigma\in S_n} \sum_{t=0}^{n-1} \log p(a^{x,\sigma}_t|c^{x,\sigma}_t ;\theta) %\\
    %\triangleq& -\mathcal L(\theta;x)
    \label{eq:jensen}
\end{align}
%We use $\mathcal L(\theta;x)$ as our loss function, which is an upper bound of $-\log p(x;\theta)$.
where equality holds when the posterior $p(\sigma | x;\theta)$ is uniform.
%Although the lower bound given by Jensen's inequality can be loose, it does not need to infer the posterior probability over trajectories (unlike variational methods) and is easy to implement.
%By maximizing this lower bound we also minimize the gap between the true posterior and the uniform approximation. As a result, we impose an inductive bias towards learning to realize $x$ equally well, independent of the order.
By maximizing this lower bound, we do not favor any particular order, but encourage the model to realize $x$ equally well in all orders.
It can help the model to complete any partial input text regardless of the position of blanks. 
%Otherwise, there may be rare intermediate states that the model does not know how to start from it.

%Since 
%\[
%\mathcal L(\theta;x)=-\log(n!) + \mathbb E_\sigma \mathbb E_t [n\cdot \log p(a^{x,\sigma}_t|c^{x,\sigma}_t ;\theta)]
%\]

A naive training algorithm is to directly estimate the lower bound in Eq.~(\ref{eq:jensen}):
%From Equation (\ref{eq:jensen}), we can derive our first naive training algorithm:
first uniformly sample a permutation $\sigma$ from $S_n$ and a step $t$ from $0$ to $n-1$, then construct the canvas $c^{x,\sigma}_t$, and compute the estimated loss $\left[-\log(n!) - n\cdot\log p(a^{x,\sigma}_t|c^{x,\sigma}_t ;\theta)\right]$.
However, this procedure has a large variance and can only compute the loss of a single action in one pass (in contrast to left-to-right language models that compute $n$ word losses per pass).

To train the model more efficiently, we note that the canvas $c^{x,\sigma}_t$ depends only on the first $t$ elements of $\sigma$.
Hence we can combine into one pass the loss calculations of trajectories that are the same in the first $t$ steps but different at the $t + 1$ step.
%$\log p(a^{x,\sigma}_t|c^{x,\sigma}_t ;\theta)$ depends only on the first $t+1$ elements of $\sigma$.
Switching the summation order of $\sigma$ and $t$, we have:
\begin{align}
    & \sum_{t=0}^{n-1} \frac{1}{n!} \sum_{\sigma\in S_n} \log p(a^{x,\sigma}_t|c^{x,\sigma}_t ;\theta) \nonumber \\
    =~ &  n\cdot\mathbb E_t\mathbb E_{\sigma_{1:t}} \mathbb E_{\sigma_{t+1}}\mathbb E_{\sigma_{t+2:n}} \left[\log p(a^{x,\sigma}_t|c^{x,\sigma}_t ;\theta)\right] \nonumber \\
    =~ &  n\cdot\mathbb E_t\mathbb E_{\sigma_{1:t}} \mathbb E_{\sigma_{t+1}} \left[\log p(a^{x,\sigma}_t|c^{x,\sigma}_t ;\theta)\right] \nonumber \\
    =~ & \mathbb E_t\mathbb E_{\sigma_{1:t}} \left[ \frac{n}{n-t} \sum_{\sigma_{t+1}} \log p(a^{x,\sigma}_t|c^{x,\sigma}_t ;\theta)\right]
\end{align}
which leads to our efficient training algorithm: sample $t$ from $0$ to $n-1$ and partial permutation $\sigma_{1:t}$, construct the canvas $c^{x,\sigma}_t$, and compute loss:
\begin{align}\label{eq:loss}
- \log(n!) - \frac{n}{n-t}\sum_{\sigma_{t+1}} \log p(a^{x,\sigma}_t|c^{x,\sigma}_t ;\theta)
\end{align}
The whole process is illustrated in Algorithm~\ref{alg:train}.
In this way, we can compute in expectation $n/2$ action losses per pass.
%\textcolor{red}{In practice, we observe that it converges much faster (how much?) than the native training algorithm.}
%We note that the Insertion Transformer can be trained in a similar way~\citep{chan2019kermit}

\section{Experiments}

\begin{figure}[t]
%\small
\fontsize{10}{12}\selectfont
\centering
  \begin{tabular}{l}
    \toprule
    %\multicolumn{1}{c}{\textsc{Text Infilling}}   \\
    %\rule{0pt}{4ex} \\
    %\hline
    %\rule{0pt}{4ex} \\
    %\midrule
    They also have \underline{\qquad} which \underline{\qquad} .\\
                     They also have \underline{\textcolor{OliveGreen}{ice cream}} which \underline{\textcolor{OliveGreen}{is really good}} . \\ 
    \midrule

    %\multicolumn{1}{c}{\textsc{Ancient Text Restoration}} \\
    %\midrule
    \texttau\textepsilon~ \textepsilon\textgamma\textgamma\textomikron\textnu\textomikron\textnu~ \textepsilon\textiota\textsigma\textalpha\textiota\underline{?\,?\,?\,?\,?\,?\,?}\textsigma\textomikron\textphi\textiota\textalpha\textiota\textvarsigma\\ 
         \texttau\textepsilon~ \textepsilon\textgamma\textgamma\textomikron\textnu\textomikron\textnu~ \textepsilon\textiota\textsigma\textalpha\textiota{\color{OliveGreen}\underline{\textomikron\textupsilon~ \,\texttau\textomikron\textupsilon ~ \,}}\textsigma\textomikron\textphi\textiota\textalpha\textiota\textvarsigma \\
            
     \midrule

    %\multicolumn{1}{c}{\textsc{Style Transfer}} \\
    %\midrule
    The employees were \textbf{{super nice}} and \textbf{{efficient}} !\\   
    The employees were \underline{\textcolor{OliveGreen}{rude}} and \underline{\textcolor{OliveGreen}{unprofessional}} !\\
    \bottomrule
    \end{tabular}
    \caption{Examples of input and output for text infilling, ancient text restoration, and style transfer tasks. % the three rewriting tasks. We contrast text infilling, where blanks can cover an arbitrary number of words, with ancient text restoration, where the number of characters to recover is indicated by the number of `?' symbols in the input.
    }
    \label{fig:example_task}
\end{figure}

We test BLM's capacity to rewrite specified portions of text on three tasks:
text infilling \citep{zhu2019textinfilling}, 
ancient text restoration \citep{assael-etal-2019-restoring}
and style transfer \citep{shen2017style}. Fig.~\ref{fig:example_task} displays example inputs and outputs for these tasks.
We also measure the perplexity of BLM on language modeling benchmarks and compare with traditional left-to-right language models.% as a sanity check. % \citep{bengio2003neural, mikolov2010recurrent, bai2018empirical-tcn}.

\begin{table*}[t]
\centering
\fontsize{9}{10.8}\selectfont
\begin{tabular}{lccccp{25pt}ccccc}
 \toprule
                      & \multicolumn{5}{c}{BLEU} & \multicolumn{5}{c}{PPL} \\
                      \cmidrule(lr{13pt}){2-6} \cmidrule(lr){7-11}
    Mask ratio        & 10\% & 20\% & 30\% & 40\% & 50\%           & 10\% & 20\% & 30\% & 40\% & 50\% \\
\midrule
No infill & 75.2 &	55.0 &	37.4 &	23.6 &	13.0 & 98.4 &	163.0 & 266.3 & 421.0 & 647.9 \\
\midrule
%InsT & 81.9 & 70.4 & 59.0 & 46.0 & 33.9 & \bf 44.7 & \bf 42.0 & 40.1 & 38.0 & 36.3 \\ had bug
InsT & 84.8 & 72.3 & 58.9 & 46.0 & 33.8 & \bf 48.3 & \bf 44.2 & 41.8 & 39.7 & 37.7 \\
MLM (oracle length) & 83.7 & 69.3 & 55.5 & 43.2 & 32.2 & 58.4 & 59.8 & 59.8 &	59.0 & 56.8 \\
BERT+LM & 82.8 & 66.3 & 50.3 & 37.4 & 26.2 & 55.1 &	55.2 & 54.9 & 56.5 & 53.6 \\
Seq2seq-full & 86.3 & 72.9 & 59.4 &	46.3 & 34.0 & 51.3	& 46.9 & 41.0 & \bf 31.9 & \bf 20.6 \\
Seq2seq-fill & 82.8 & 67.5 & 52.9 &	39.9 & 28.6 & 64.6 &	71.0 & 73.4 & 65.6 & 48.7 \\
\midrule
BLM & \bf 86.5 & \bf 73.2 & \bf	59.6 & \bf 46.8 & \bf 34.8 & 50.2 & 44.9 & \bf 39.9 & 35.0 & 32.7\\
 \bottomrule

\end{tabular}
\caption{BLEU scores and perplexity of generated documents by different models for text infilling. The perplexity is measured by a pre-trained left-to-right language model, and the original documents have perplexity 55.8.
%The ``No infill'' line reports the BLEU score of the blanked document.
}\label{tab:infill_results}
\end{table*}

\begin{table}[t]
\centering
\fontsize{9}{10.8}\selectfont
\begin{tabular}{lccccc}
 \toprule
Mask ratio & 10\% & 20\% & 30\% & 40\% & 50\% \\
\midrule
Seq2seq-full& 15.0 & 22.4 &	28.7 & 33.3 & 40.6 \\
Seq2seq-fill& 31.0 & 28.4 &	34.5 & 42.5 & 47.2 \\
 \bottomrule
\end{tabular}
\caption{Infilling failure rate (\%) of seq2seq models. Other methods always produce valid outputs.}\label{tab:infill_err_rate}
\end{table}

%\textbf{Experimental Details}\quad
\paragraph{Experimental Details}
In all experiments, the sequence representations in BLM are obtained using
the encoder module of \texttt{transformer\_base} \citep{vaswani2017-attention} ($6$ layers, $8$ heads, $d_{model}=512$, $d_{ff}=2048$, $d_k=d_v=64$).
The MLP used for blank prediction has one hidden layer of size $1024$.
Weight decay, learning rate, and dropout are tuned based on the loss on the validation set for each dataset respectively. 
When decoding, we use beam size in $\{1, 5, 10\}$ and choose the best value as observed on the validation set.
We note that beam search in BLM does not search for the sentence with the maximum marginal likelihood $p(x; \theta)$, but instead for a sentence \emph{and} a trajectory that have the maximum joint likelihood $p(x, \sigma; \theta)$.

\subsection{Text Infilling}

\iffalse
\begin{figure*}[t]
\centering
\includegraphics[width=\textwidth]{figs/infill.pdf}
\caption{Failure rate, BLEU score and perplexity of generated documents for the text infilling task.
The ``No infill'' line reports the BLEU score of the blanked document.
The ``Data PPL'' dotted line serves as reference for the perplexity of the original documents}\label{fig:infill_results}
\end{figure*}
\fi

\iffalse
The task of text infilling is motivated by many practical applications where the goal is to augment partially completed documents with missing information~\citep{zhu2019textinfilling}.
Following the protocol of \citet{zhu2019textinfilling}, we automatically compile test data by deleting portions of documents, and ask systems to fill them in.
%where partially completed documents are created by deleting portions of the text, and ask systems to fill in. 
The first row in Fig.~\ref{fig:example_task} showcases an example input-output pair.
%As such, text infilling is the ideal task to evaluate the BLM's ability
The infilling task evaluates model's ability
to complete blanks in a document while maintaining semantic consistency with the imposed context.
\fi

\paragraph{Dataset}
We experiment on the Yahoo Answers dataset, which has 100K/10K/10K documents for train/valid/test respectively \citep{yang2017improved}.
A document has a maximum length of 200 words, with an average of 78 words.
Following \citet{zhu2019textinfilling}, we automatically compile test data by deleting portions of documents.
For each document $x$, we randomly mask a given ratio $r$ of its tokens. 
Contiguous masked tokens are collapsed into a single ``\underline{\quad}'',
resulting in a canvas $c$ to be completed. % with $k$ such blanks.
%Systems are required to complete the blanks in $c$.
%, creating a sequence $x$ of tokens, $k$ of which are blanks. 

%The text infilling task involves reconstructing the original sentence $s$ from the blanked sequence $x$ or, equivalently, recovering the sequence $(y_1, \dots, y_k)$ of blanked tokens, where each $y_i$ corresponds to a blank in $x$ and is comprised of one or multiple words.
%In the previous example, $k=2$ and $(y_1, y_2) = ($ice cream $,$ is really good$)$.

\paragraph{Metrics}
%Following prior work \citep{zhu2019textinfilling, liu2019-tigs}, 
We measure generation's accuracy by computing its BLEU score against the original document $x$, and fluency as its perplexity evaluated by a pre-trained (left-to-right) language model.
We also report the failure rate, which is the percentage of invalid generations, such as missing existing words or not filling in all the blanks.

\paragraph{Baselines}
We compare BLM with five baselines:
\begin{itemize}[topsep=3pt]
\setlength\itemsep{0pt}
\item \emph{Insertion Transformer (InsT)}: % is trained the same way as the BLM, by randomly sampling a canvas at each training step.
\iffalse
Since InsT does not support controlling the insertion position,
we force it to generate words only in the designated blanks by normalizing the predictions over valid locations. 
We further mask the $\langle$eos$\rangle$ prediction unless all blanks have been filled with at least one token. However, sometimes when the maximum length is reached, the generation is not yet complete. Hence, we prioritize the slots that have not been filled yet when choosing which slot to insert in.
\fi
By default, InsT does not support controlling the insertion position. We force it to produce valid generations by normalizing the predictions over valid locations, disabling the $\langle$eos$\rangle$ prediction unless all blanks have been filled, and prioritizing slots that have not been filled yet.
Without these steps, InsT would have a failure rate $\ge88\%$.

\item \emph{MLM (oracle length)}: MLM for text infilling requires predicting the length of each blank. Here we replace blanks with the target number of $\langle$mask$\rangle$ tokens, and fill them autoregressively by the most-confident-first heuristic. 

\item \emph{BERT+LM}: We use BERT’s representation of each blank as seed for a left-to-right language model that learns to generate the tokens in the corresponding blank.
At inference time, the multiple blanks are filled in one after another, conditioned on previous generations.

\item \emph{Seq2seq-full} \citep{donahue2020enabling}: We train a seq2seq model to output the full document $x$ from input $c$. Note that it may have invalid outputs that do not match the input format, such as missing existing tokens in $c$ or generating tokens in incorrect locations.
% copying real tokens from $x$ and replacing blank tokens when necessary.
% e.g. ``They also have ice cream which is really good''.
\item \emph{Seq2seq-fill} \citep{donahue2020enabling}: We train a seq2seq model to output only tokens to be placed in the blanks, with a special `\textbar' token to indicate separation. For the example in Fig.~\ref{fig:example_task}, its target output will be ``ice cream \textbar is really good''. Unlike \emph{seq2seq-full}, \emph{seq2seq-fill} does not have the problem of losing existing tokens in $c$. However, it may still fail to generate the correct number of `\textbar' that matches the input.

%missing tokens $(y_1,\dots, y_k)$. 
%In practice, the model is trained to predict the string $y_1$\texttt{<sep>}$\dots$\texttt{<sep>}$y_k$, where \texttt{<sep>} is a special token that separates two blanks, e.g. ``ice cream \texttt{<sep>} is really good''.

\end{itemize}

\begin{figure*}[t]
\fontsize{8}{9.6}\selectfont
\centering
    % TODO later: prettier underlines
% Prettier table separation
\begin{tabular}{p{37pt}p{180pt}p{200pt}}
\toprule

& \multicolumn{1}{c}{Mask-ratio 10\%} & \multicolumn{1}{c}{Mask-ratio 50\%} \\
\midrule
Blanked & 
when time flies , \underline{\qquad} does it go ? \underline{\qquad} the center of the \underline{\qquad} to be recycled \underline{\qquad} made into new time .
&
when time \underline{\qquad} , where \underline{\qquad} ? \underline{\qquad} the \underline{\qquad} of \underline{\qquad} universe to \underline{\qquad} recycled \underline{\qquad} made into \underline{\qquad} .
\\
\midrule
BLM &
when time flies , \underline{\emph{where}} does it go ? \underline{\emph{for}} the center of the \underline{\emph{earth}} to be recycled \underline{\emph{and}} made into new time .\underline{\emph{}}
& 
\underline{\emph{}}when time \underline{\emph{was created}} , where \underline{\emph{did it come from}} ? \underline{\emph{it was}} the \underline{\emph{first part}} of \underline{\emph{the}} universe to \underline{\emph{be}} recycled \underline{\emph{and}} made into \underline{\emph{space}} .\underline{\emph{}}
\\[7pt]
InsT 
& 
\underline{\emph{}}when time flies , \underline{\emph{where}} does it go ? \underline{\emph{for}} the center of the \underline{\emph{earth has}} to be recycled \underline{\emph{and}} made into new time .
&
when time \underline{\emph{was created}} , where \underline{\emph{was it}} ? \underline{\emph{what was}} the \underline{\emph{name}} of \underline{\emph{the}} universe to \underline{\emph{be}} recycled \underline{\emph{and}} made into \underline{\emph{space}} .
\\[7pt]
MLM ~~~~~~~(oracle len) &
\underline{\emph{}}when time flies , \underline{\emph{where}} does it go ? \underline{\emph{from}} the center of the \underline{\emph{earth}} to be recycled \underline{\emph{converted}} made into new time .\underline{\emph{}}
& 
\underline{\emph{}}when time \underline{\emph{is}} , where \underline{\emph{is the universe}} ? \underline{\emph{from}} the \underline{\emph{creation}} of \underline{\emph{the}} universe to \underline{\emph{be}} recycled \underline{\emph{and}} made into \underline{\emph{the universe}} .\underline{\emph{}}
\\[7pt]
BERT+LM &
\underline{\emph{}}when time flies , \underline{\emph{where}} does it go ? \underline{\emph{to}} the center of the \underline{\emph{earth}} to be recycled \underline{\emph{came}} made into new time .\underline{\emph{}}
& 
\underline{\emph{}}when time \underline{\emph{is}} , where \underline{\emph{to}} ? \underline{\emph{i need to find}} the \underline{\emph{way}} of \underline{\emph{the}} universe to \underline{\emph{be}} recycled \underline{\emph{and}} made into \underline{\emph{a lot}} .\underline{\emph{}}
\\[7pt]
Seq2seq-full &
\underline{\emph{}}when time flies , \underline{\emph{where}} does it go ? \underline{\emph{at}} the center of the \underline{\emph{earth}} to be recycled \underline{\emph{and}} made into new time .\underline{\emph{}}
&
\underline{\emph{}}when time \underline{\emph{heals}} , where \underline{\emph{does it go}} ? \underline{\emph{it 's}} the \underline{\emph{end}} of \underline{\emph{the}} universe to \underline{\emph{be}} recycled \underline{\emph{and}} made into \underline{\emph{space}} .\underline{\emph{}}
\\[7pt]
\multirow{2}{30pt}{Seq2seq-fill}
&\underline{\emph{}}when time flies , \underline{\emph{how}} does it go ? \underline{\emph{at}} the center of the \underline{\emph{earth}} to be recycled \underline{\emph{and}} made into new time .\underline{\emph{}}	
& 
\underline{\emph{}}when time \underline{\emph{is time}} , where \underline{\emph{is time}} ? \underline{\emph{time is}} the \underline{\emph{time}} of \underline{\emph{time}} universe to \underline{\emph{the}} recycled \underline{\emph{be}} made into \underline{\emph{and}} .\underline{\emph{}} \textcolor{red}{\emph{the universe}}
\\
&
\multicolumn{1}{l}{how \textbar at \textbar earth \textbar and} &
is time \textbar is time \textbar time is \textbar time \textbar time \textbar the \textbar be \textbar and \textbar the universe
\\
\midrule
Original & 
\textbf{}when time flies , \textbf{where} does it go ? \textbf{to} the center of the \textbf{universe} to be recycled \textbf{and} made into new time .	
&
\textbf{}when time \textbf{flies} , where \textbf{does it go} ? \textbf{to} the \textbf{center} of \textbf{the} universe to \textbf{be} recycled \textbf{and} made into \textbf{new time} .\\
\bottomrule
\end{tabular}

\caption{Example generations of different models for text infilling on Yahoo Answers. %, with mask ratios 0.1 and 0.5.
Completions are in italic.
Invalid completions are in red.
For Seq2seq-fill, we present model outputs along with the merged document.
%In this example,  the insertion transformer produces invalid completions by failing to generate tokens in the ``? \underline{\quad} the''  blank. 
%At mask ratio 50\%, the \emph{seq2seq-fill} baseline also generates an invalid document by producing too many `\textbar' tokens, i.e. filling too many blanks.
}
\label{fig:infill_gens}
\end{figure*}

\paragraph{Results}

%In Figure~\ref{fig:infill_results}, we plot the failure rate, BLEU score, and perplexity of models at different mask ratios.
%Our BLM is the only method that is able to consistently generate valid outputs.
%Insertion Transformer has the highest failure rate: in more than $88\%$ of cases, it does not fill all the blanks.
%This indicates that the Insertion Transformer is not suitable for generation with location constraints.

%On the lower end, at $r=0.1$ the \emph{seq2seq-full} and \emph{seq2seq-fill}  baselines have failure rate 15\% and 32\% respectively. 
%For $r=0.5$ the failure rate is bumped to 41\% and 56\%.
%Surprisingly, the Insertion Transformer fails to produce valid completions for a very high number of documents.

%The examination of the generated outputs reveals that the insertion transformer chooses to not complete certain required blanks. It can be attributed to the Insertion Transformer's formulation which does not distinguish between initial blanks from blanks left open after filling in a first word. 

As shown in Table~\ref{tab:infill_results}, BLM achieves the highest BLEU score at all mask ratios: 0.7 to 1.7 higher than InsT, 2.6 to 4.1 higher than MLM with oracle length, and 3.7 to 9.4 higher than BERT+LM.
InsT is not trained with insertion position control. Restricting it to generate at the specified positions %falls short of BLM's performance.
%InsT falls short of BLM's performance, since it is not trained with insertion position control. To ensure valid generations, we have to impose constraints on InsT at inference time, which
thus biases the model towards making suboptimal completions.
MLM is trained to independently predict masked tokens instead of jointly modeling them.
Even with the target number of $\langle$mask$\rangle$ tokens given, its performance is still inferior to BLM.
BERT+LM lags behind other models. In BERT training, one mask corresponds to one token, whereas a blank here can cover multiple tokens, and the distance between words is not fixed. Hence, it is difficult for the LM to complete the sentence from BERT representations.

Seq2seq-full has BLEU scores closest to BLM. However, its failure rate ranges from $15\%$ to $40.6\%$ as the mask ratio increases.
Seq2seq-fill performs worse than Seq2seq-full, possibly because the decoder has to model segmented text while counting the number of blanks.
%According to the BLEU score, BLM and \emph{seq2seq-full} have the highest infilling accuracy, on average $5.8$ points higher than that of the Insertion Transformer and \emph{seq2seq-fill}.
%The BLM also outperforms all other baselines in terms of accuracy, as measured by the BLEU score.
%For reference, we also plot the BLEU score of the input canvas with respect to the original document. 
%When the mask ratio is $0.5$, the input BLEU score is $13.0$, and BLM brings it up to $34.8$ after infilling.
%We observe that, as the mask ratio increases, the gain in BLEU score over the blanked document remains stable or increases.

In terms of fluency, outputs of BLM, InsT and Seq2seq-full all have perplexity lower than original data perplexity. 
This is because with beam search, models tend to generate the most typical output with the highest likelihood~\citep{holtzman2019curious}.

%are able to bring down the perplexity of the document\footnote{Perplexities of the blanked documents are respectively 98.36, 163.01, 266.32, 420.97 and 647.92} to around 56, the perplexity of the target, as evaluated by the same pretrained language model. 
%The Insertion Transformer's perplexity is closer to the target's than the BLM's.
%Indeed, by preferring to not complete certain required blanks, the insertion transformer tends to omit words that carry less entropy (e.g. ``I believe'' instead of ``I also believe''), yielding a higher perplexity.

Examination of model generations confirms the superiority of BLM.
In Fig.~\ref{fig:infill_gens}, we showcase example outputs by each model at different mask ratios.
In low mask ratio settings, models only need to fill in the blanks with a single word to produce grammatical completions. Most models succeed in this task.
With a higher mask ratio of 50\%, %where half of the words are deleted and
the main ideas of the document are concealed, and the infilling task is much more challenging. Models need to creatively generate sentences that fit the imposed canvas.
Although the original meaning of the sentence is not recovered, BLM is the only model able to produce a coherent document with consistency between the question and the answer.

Overall, BLM displays the best performance both quantitatively and qualitatively.
Its inherent text infilling ability frees it from length, order, or termination heuristics used by other methods.
\subsection{Ancient Text Restoration}
Ancient text restoration is a form of text infilling where there are fragments in ancient documents that are illegible due to time-related damages and need to be recovered. \citet{assael-etal-2019-restoring} introduces the PHI-ML dataset made of fragments of ancient Greek inscriptions.
%The task of ancient text restoration requires recovering missing portions of text in a document~\citep{assael-etal-2019-restoring}. 
%This capacity is essential for analyzing ancient documents with illegible fragments due to time-related damages.
%The second row in Fig. \ref{fig:example_task} illustrates an example of input and output for the task.
Restoration is performed at the character-level. The number of characters to recover is assumed to be known and indicated by a corresponding number of `?' symbols, as shown in the second row of Fig.~\ref{fig:example_task}.
In reality, when epigraphists restore a deteriorated document, the length of the lost fragment is unknown and needs to be guessed as a first step. 
While models proposed by \citet{assael-etal-2019-restoring} relies on expert conjectures, we note that BLM can bypass this limitation and flexibly generate completions without this additional knowledge.
However, in order to compute the character error rate (CER) for each '?' and have a fair comparison with previous work, we evaluate our model in the length-aware setting.

\paragraph{Length-aware BLM (L-BLM)}
We present a variant of BLM adapted to the specific features of this task. The vocabulary $V$ is an alphabet of characters from the ancient Greek language. We extend $V$ with special ``\underline{\quad\texttt{[}$t$\texttt{]}\quad}'' tokens that denote the length of the fragment to recover. Specifically, as a preprocessing step, consecutive `?' characters are collapsed into a single ``\underline{\quad\texttt{[}$t$\texttt{]}\quad}'' token, where $t$ is the number of `?' symbols.
For each such blank token, L-BLM is trained to predict a character to fill in and the length $l\in\{0,\cdots,t-1\}$ of the new blank to its left. The length of the new blank on the right is accordingly $t-1-l$.
%In all experiments, we use special blank tokens for lengths up to $1000$ and follow our usual canvas creation procedure. 

\paragraph{Dataset} 
The PHI-ML dataset contains about 3 million words / 18 million characters.
We evaluate models in two settings: \emph{single-slot} and \emph{multi-slot}.
%The test set is generated following \citealt{assael-etal-2019-restoring}'s procedure:
For the single-slot setting, we use the testing script of \citet{assael-etal-2019-restoring} which samples
a context of length $L=1000$ from an inscription, then samples a slot of length $C\in [1, 10]$ from that context.
The characters from the slot are replaced with `?' and constitute the target.
% In both cases, to ensure compatibility with the baseline considered, we use $L=1000$ and sample slots of length $C\in [1, 10]$.
%For the single-slot experiment, we use the testing script from \citet{assael-etal-2019-restoring} and sample 12,800 testing samples, for a total of 63,234 characters to predict, with mask ratio of 1.2\%.
For the multi-slot setting, we progressively increase the number of slots, yielding mask ratios of 25\%, 40\% and 50\% respectively.
%In total, we generate a total of 1000 samples for each mask ratio of 25\%, 40\% and 50\% with respectively 150,235, 400,827 and 406,231 characters to restore.
% The random seed is fixed to evaluate all models against the same blanked inscriptions.

%  \begin{figure}[t]
%  \centering
% \fontsize{9}{10.8}\selectfont
% \begin{tabular}{cccc}
%  \toprule
%   Split & Inscriptions & Word & Chars \\
%  \midrule
%  Train & 34,952 & 2,792K & 16,300K \\
%  Dev & 2,826 & 211K & 1,230K \\
%  Test & 2,949 & 223K & 1,298K \\
%  \bottomrule
% \end{tabular}
% \caption{Statistics of the PHI-ML dataset.}\label{fig:ancient_text_dataset}
% \end{figure}

\paragraph{Baselines}
\citet{assael-etal-2019-restoring} proposed two models:
\emph{Pythia}, a character-level seq2seq-based approach; %specialized in ancient text restoration
and
\emph{Pythia-Word}, a variant of Pythia that uses both character and word representations as input.
During training, the model learns to recover the missing characters of examples where a random slot has been masked. %, with a slot length limited to 10.
When testing on the multi-slot setting, Pythia(-Word) is applied iteratively with beam size 20 for each slot. %as described in \citealt{assael-etal-2019-restoring}.

%\paragraph{Metric}
%We measure the character error rate (CER) of all models in both settings.

\begin{table}[t]
\centering
\fontsize{9}{10.8}\selectfont
\begin{tabular}{lcccc}
 \toprule
                      & Single- & \multicolumn{3}{c}{Multi-slot} \\
                      \cmidrule(lr){2-2} \cmidrule(lr){3-5}
    Mask ratio        & 1\%           & 25\% & 40\% & 50\% \\
 \midrule
 Human       & 57.3\%         &    -            &      -          &  -  \\
Pythia       & 32.5\%         &    -            &      -          &  -  \\
Pythia-Word & \textbf{29.1\%}& \textbf{36.9\%}&   42.3\%         & 44.9\% \\
\midrule
L-BLM               & 33.7\%         &    37.1\%      & \textbf{37.9\%} &  \textbf{41.6\%}     \\
 \bottomrule

\end{tabular}
\caption{CER for ancient text restoration.}\label{tab:ancient_text_results}
\end{table}

\paragraph{Results}
Table \ref{tab:ancient_text_results} summarizes the CER of all models in both settings.
L-BLM achieves similar CER as Pythia in the single-slot setting, significantly outperforming human experts.
Augmented with word representations, Pythia-Word further decreases the error rate compared to character-only methods.

%[resolved] RB Their baseline does better even in 25%. I am not sure if your description gives them sufficient credit.
In reality, restoring damaged inscriptions requires reconstructing multiple lost fragments.
As a larger proportion of text is missing, Pythia-Word's performance is degraded.
L-BLM is robust to the setting change
and outperforms Pythia-Word at the mask ratio of 40\% and 50\% by 4.4 and 3.3 points, respectively.
We posit that L-BLM's advantage lies in its ability to maximize the joint likelihood of the completions over all slots. 
In contrast, Pythia-Word's is only aware of one slot at a time, and beam search is performed locally within each slot.
%thus suffers from suboptimal inference.
%Moreover, L-BLM can handle slots of arbitrary long length while Pythia-Word is limited to slots of up to 10 characters, which is a limiting factor for real-world usage.
\subsection{Sentiment Transfer}

The goal of sentiment transfer is to modify the sentiment of a sentence while maintaining its topic \citep{shen2017style}. An example is described on the third row of Fig. \ref{fig:example_task}.
Inspired by the way humans perform rewriting, we follow a recent line of work in style transfer that adopts a two-step approach \citep{li-etal-2018-delete, xu2018unpaired, wu2019maskinfill}:

\begin{enumerate}[topsep=3pt]
\setlength\itemsep{0pt}
    \item Remove words and expressions of high polarity from the source sentence;
    \item Complete the partial sentence with words and expressions of the target sentiment.
\end{enumerate}

Specifically, we adapt the \emph{Mask-And-Infill (M\&I)} framework of \citet{wu2019maskinfill}.
%VQ we show improvements over previous approaches for both of them
%Step 1 has been performed in previous work by masking tokens either based on their frequency-ratio \citep{li-etal-2018-delete, wu2019maskinfill} or their attention scores \citep{xu2018unpaired, wu2019maskinfill}.
We perform Step~1 by training a Bi-LSTM sentiment classifier and masking words whose attention weight is above average.
% \textcolor{red}{Step 2 is usually achieved by optimizing the reconstruction loss of a generator model conditioning on a sentiment.} 
%Step 2 is performed by various sequence models conditioning on the masked sentence and the target sentiment.
We evaluate the contribution of our model as an infilling module in Step 2 in place of their fine-tuned BERT model.
 %\citep{li-etal-2018-delete,wu2019maskinfill}.
To this end, we train two instances of BLM on the dataset, one for each sentiment. 
At test time, the corresponding BLM is used to produce completions of the target sentiment.

\citet{wu2019maskinfill} further train the infilling model with the classifier to improve transfer accuracy.
They use soft words relaxation to backprop gradients from the classifier to the generator.
For BLM, however, we cannot pick locations or insert blanks as ``soft'' choices, making it challenging to employ a classifier at training time.
Nevertheless, we can easily apply the classifier to guide inference. We sample 10 outputs and keep the one with the highest classifier ranking. It is not slower than beam search with size 10 and can be fully parallelized.

\paragraph{Datasets} We test on the Yelp and Amazon review datasets \citep{shen2017style,li-etal-2018-delete}. 
The Yelp dataset has 450K/4K/1K non-parallel sentences for train/valid/test respectively, and
the Amazon dataset has 555K/2K/1K sentences.
%(see Table \ref{fig:style_transfer_dataset}).
Each sentence is labeled as either positive or negative. 
%The task is to flip the sentiment of each sentence.

%  \begin{figure}[t]
%  \centering
% \fontsize{9}{10.8}\selectfont
% \begin{tabular}{ccccc}
%  \toprule
%   Attribute & Train & Dev & Test \\
%  \midrule
%  Positive & 270K & 2000 & 500 \\
%  Negative & 180K & 2000 & 500 \\
%  \bottomrule
% \end{tabular}
% \caption{Statistics of the Yelp review dataset for style transfer.}\label{fig:style_transfer_dataset}
% \end{figure}

\paragraph{Metrics}
We use evaluation methods introduced by prior work \citep{shen2017style, li-etal-2018-delete}. %, wu2019maskinfill, yang2018unsupervised}.
To assess the accuracy of generated sentences with respect to the target sentiment, we use a pretrained CNN classifier that achieves 97.7\%  accuracy on the Yelp dataset and 82.2\% accuracy on the Amazon dataset.
We also measure the BLEU score between transferred sentences and human references. %\citep{li-etal-2018-delete}. 

\begin{table}[t]
\centering
\fontsize{9}{10.8}\selectfont
\begin{tabular}{lcccc}
\toprule
& \multicolumn{2}{c}{Yelp} & \multicolumn{2}{c}{Amazon} \\
\cmidrule(lr){2-3} \cmidrule(lr){4-5}
& ACC & BLEU & ACC & BLEU \\
\midrule
\citet{li-etal-2018-delete} & 88.7 & 8.4 & 48.0 & 22.8 \\
%NAE & 80.0 & 22.5 & 70.4 & 14.1 \\
\citet{zhang2018style} & 96.6 & 22.8 & 84.1 & 33.9 \\
\citet{wu2019hierarchical} & 91.5 & \bf 29.9 & 40.2 & \bf 41.9 \\
M\&I with MLM & 41.5 & 15.9 & 31.2 & 32.1 \\
~~ + classifier & \bf 97.3 & 14.1 & 75.9 & 28.5 \\
\midrule
%Mask only (canvas) & 42.6 & 19.9 & 37.1	& 21.2 \\
M\&I with BLM & 79.6 & 21.9 & 52.0 & 24.7 \\
~~ + classifier & 96.5 & 21.5 & \bf 92.5 & 23.1 \\
\bottomrule
\end{tabular}
\caption{Accuracy and BLEU scores for style transfer. %Accuracy measures the percentage of sentences labeled as the target sentiment by the classifier. BLEU is evaluated against human reference generations. For reference, we also report accuracy and BLEU scores of the canvas (i.e. the original masked sentence).
}\label{tab:style_transfer_results}
\end{table}

\begin{figure}[t]
\centering
\fontsize{7.9}{9.5}\selectfont
\begin{tabular}{p{207pt}}
%\begin{tabular}{l}
 \toprule
 
%\textbf{}the food 's ok , the service is \textbf{among} the \textbf{worst} i have encountered .\textbf{} \\
%\underline{\emph{}}the food 's ok , the service is \underline{\emph{probably}} the \underline{\emph{best}} i have encountered .\underline{\emph{}} \\
%the food is good, and the service is one of the best i've ever encountered.\\
%\midrule

\textbf{}everyone that i spoke with was \textbf{very helpful} and \textbf{kind} .\textbf{} \\
\underline{\emph{}}everyone that i spoke with was \underline{\emph{rude}} and \underline{\emph{unprofessional}} .\underline{\emph{}} \\
everyone that i spoke with wasn't helpful or kind.\\
\midrule

\textbf{}the beans were in the burro in the rice was \textbf{nowhere to be} found .\textbf{} \\
\underline{\emph{}}the beans were in the burro in the rice was \underline{\emph{the best i}} found .\underline{\emph{}} \\
the beans were in the burro and the rice was plentiful\\
\midrule

%\textbf{}everything is \textbf{fresh} and so \textbf{delicious} !\textbf{} \\
%\underline{\emph{}}everything is \underline{\emph{horrible}} and so \underline{\emph{expensive}} !\underline{\emph{}} \\
%everything was so stale\\
%\midrule

\textbf{}there is \textbf{definitely not} enough \textbf{room} in that part of the venue .\textbf{} \\
\underline{\emph{}}there is \underline{\emph{always}} enough \underline{\emph{parking}} in that part of the venue .\underline{\emph{}} \\
there is so much room in that part of the venue\\
\midrule

\textbf{}it is n't \textbf{terrible} , but it is \textbf{n't} very good either .\textbf{} \\
\underline{\emph{}}it is n't \underline{\emph{fancy}} , but it is \underline{\emph{still}} very good either .\underline{\emph{}} \\
it is n't perfect , but it is very good .\\
%\midrule

%\textbf{}executive chefs would \textbf{walk} by \textbf{not} even saying good morning .\textbf{} \\
%\underline{\emph{}}executive chefs would \underline{\emph{come}} by \underline{\emph{without}} even saying good morning .\underline{\emph{}} \\
%the excecutive chef was nice and said good morning to us very often\\

\bottomrule
\end{tabular}
\caption{Example generations by BLM for sentiment transfer on Yelp. The first line is the source sentence with masked words in bold. The second line is BLM's completion. The third line is a human reference. 
% task using attention-based masking mechanism.
}\label{fig:style_gens}
\end{figure}

\paragraph{Results}
In Table~\ref{tab:style_transfer_results}, we can see that directly applying BLM as the infilling module is significantly better than MLM. The accuracy on Yelp and Amazon datasets is increased by 38.1\% and 20.8\%, respectively.
In addition to the aforementioned problem of MLM being trained to predict masked tokens independently, it must generate the same number of tokens as in the source sentence, whereas our BLM formulation is not subject to this constraint.
Our simple use of a classifier at inference time further improves accuracy. It achieves the highest accuracy of 92.5\% on Amazon with a small decrease in BLEU, indicating that BLM can easily find high-quality outputs.

%Results in Table \ref{tab:style_transfer_results} demonstrate the ability of different models to perform text infilling for style transfer. 
%The \textsc{Delete-And-Retrieve} method with the frequency-ratio based masking strategy achieves high sentiment accuracy, but can only do so at the expense of content fidelity.
%By constraining BLM to fill in blanks in between content words, we ensure that the predictions will yield high content preservation, improving both BLEU score and sentiment accuracy over the original masked sentence.

%The MLM formulation in \textsc{Mask-And-Infill} is problematic on this task for two reasons.
%By design, MLM is forced to generate the same number of tokens as there were originally in the source sentence, making it more difficult to produce coherent sentences that are consistent with the target sentiment.
%Furthermore, MLM is trained to predict the masked tokens independently rather than jointly, which further hurts performance.
%Our formulation of BLM does not suffer any of these weaknesses.
%With both masking strategies, our model outperforms the \textsc{Mask-And-Infill} baseline on all metrics, proving its superiority as the better-suited formulation for this setup.

In Fig.~\ref{fig:style_gens}, we show examples generated by BLM on Yelp. 
It can dynamically adapt to the imposed canvas and fill in blanks with expressions of varied lengths, e.g., ``\textbf{nowhere to be} found'' $\rightarrow$ ``\underline{the best i} found'' and ``\textbf{definitely not}'' $\rightarrow$  ``\underline{always}''.
We note that failure cases arise when negative words like ``either'' are left unmasked; BLM is then unable to produce satisfactory outputs from the canvas.

\subsection{Language Modeling}
% Our final experiments study the perplexity of BLM in comparison with left-to-right language models. Since BLM is designed  for the infiltering task, it is not optimized for stndard LM benchmarks. At the same time, by analyzing its performance on these benchmarks gives us additional insight on its properties.
Language modeling is a special case of text infilling where sequences are generated from scratch.
Traditional left-to-right models dominate this task, but are not suitable for text infilling. Conversely, unconventional sequence models are rarely evaluated on language modeling. Here, we study the perplexity of BLM and Insertion Transformer, and compare them with left-to-right language models to provide additional insights.% on its properties.
%To compute the perplexity of BLM% and the Insertion Transformer

We use the Monte-Carlo method to estimate the likelihood in Eq.~(\ref{eq:likelihood}) with $m$ samples.
While the estimate is unbiased, given that per-word perplexity is a convex function of per-sentence likelihood, sampling estimates like ours are likely yielding a value higher than the actual perplexity (see Appendix~\ref{sec:mc} for a proof). 
As $m$ increases, it converges to the actual perplexity.
%With a larger $m$, the estimated perplexity is smaller and closer to the real perplexity
%We choose $m$ that we observe close to convergence on the dataset.

% \textbf{Datasets}\quad
\paragraph {Datasets}
We test on three benchmark datasets: Penn Treebank (PTB) which has about 1M tokens \citep{mikolov2010recurrent}, WikiText-2 (WT2) which has 2M tokens, and WikiText-103 (WT103) which has 103M tokens \citep{merity2016pointer}.
% \textbf{Perplexity Estimation}\quad

\begin{table}[t]
\centering
\fontsize{9}{10.8}\selectfont
\begin{tabular}{ccccc}
 \toprule
$m$ & 1 & 10 & 100 & 1000\\
\midrule
Estimated PPL 	
& 46.3 & 44.4 & 43.3 & 42.5 \\
 \bottomrule
\end{tabular}
\caption{The estimated perplexity of BLM with the number of MC samples $m$ on WikiText-103.}\label{tab:lm_mc}
\end{table}

\begin{table}[t]
\centering
\fontsize{8.35}{10}\selectfont
\begin{tabular}{lccc}
 \toprule
               & PTB   \hspace{-5pt}  & WT2 \hspace{-5pt} & WT103\\
 \midrule
LSTM  \citep{grave2016improving-lstm}         &  82.3  &   99.3 &      48.7        \\
AWD-LSTM \citep{merity2017regularizing} & 57.3 & \textbf{65.8} & - \\
TCN \citep{bai2018empirical-tcn}            &  88.7  &   -     &  45.2   \\
Transformer \citep{dai2019transformer-xl}    &  -  &    -       &   30.1    \\
% Adaptive Input \citep{baevski2018adaptive} \hspace{-15pt}   &  -  &    -       &   18.7    \\
Adaptive \citep{baevski2018adaptive}   &  -  &    -       &   18.7    \\
Transformer-XL \citep{dai2019transformer-xl} &  \textbf{54.5}  &    -       &  \textbf{18.3}         \\
\midrule
InsT (our implementation)     &  77.3  & 91.4 & 39.4    \\
BLM            & 69.2   & 81.2 & 42.5\\
 \bottomrule
\end{tabular}
\caption{Perplexity on the PTB and WikiText datasets.}\label{tab:lm_results}
\end{table}

\paragraph{Results}
Table~\ref{tab:lm_mc} shows the trend of estimated PPL with the number of samples $m$. We choose $m=1000$ in our evaluation, which is close to convergence.
Table~\ref{tab:lm_results} summarizes the perplexity of our model in comparison with previous work. 
The top results are achieved by the Transformer-XL \citep{dai2019transformer-xl} and the adaptive embedding method \citep{baevski2018adaptive}. 
%However, these systems have additional advantages in terms of utilization of supplementary techniques and their model size. On WikiText-103, the models in \citet{dai2019transformer-xl} and \citet{baevski2018adaptive} use 250M parameters, whereas our model uses 42M parameters.
They use larger model sizes and supplementary techniques that can also be combined with our model.
%When evaluated against comparable baselines,
BLM rivals the Insertion Transformer and outperforms left-to-right language models with LSTM and Temporal Convolutional Network (TCN) architecture.
%This is particularly noteworthy, since the language modeling task is more challenging for free-order models like ours.
Language modeling seems to still be challenging for free-order models. By reporting the perplexity of unconventional models like BLM, we hope to stimulate future work in this area to close the performance gap with traditional left-to-right models.

% we believe these models have the potential to close the performance gap with traditional left-to-right models.
% We hope that by reporting the perplexity of unconventional models like BLM, we can inspire future work to explore the potential of free-order language models for text generation.

\iffalse
Table~\ref{tab:lm_results} shows that our proposed BLM compares favorably against the Insertion Transformer as well as left-to-right language models with LSTM and Temporal Convolutional Network (TCN) architecture. 

We emphasize that we are not to compete for the state-of-the-art perplexity achieved by larger models such as the Transformer-XL \citep{dai2019transformer-xl} or models using additional techniques like adaptive softmax~\citep{baevski2018adaptive}.
Rather, we are the first to report decent perplexity of generalized sequence models that accommodate different generation orders. \textcolor{red}{change english}
\fi
%it validates the potential of the blank language model for sequence modeling tasks.

\section{Conclusion}
% meta-review
% This paper proposes a text generation model specifically for text-infilling. The model consumes a canvas with one or more blanks, and parameterizes an action distribution over which blank to fill, and whether to fill it with just a word, a word preceded by a blank, a word followed by a blank, or both. This model distinguishes itself from other common text-infilling models by explicitly modeling which positions are to be filled in and which to be left intact. The authors train by optimizing a lower bound on the marginal likelihood of a sequence (which sums over all possible derivations). The authors experiment on 3 tasks and report results that either improve or are competitive with reasonable baselines.
% The reviewers generally appreciated the model and the experimental results. R3 was concerned about the proposed approach's difference from the Insertion Transformer, though I believe this concern was largely addressed in the author response.

In this paper, we proposed the Blank Language Model for flexible text generation.
%BLM can generate sequences in different orders by dynamically creating and filling in blanks.
Given partially specified text with one or more blanks, BLM will fill in the blanks with a variable number of tokens consistent with the context.
We demonstrate the effectiveness of our model on various text rewriting tasks, including text infilling, ancient text restoration and style transfer.

% Re R2
% We thank the reviewer for the feedback. Checking BLM’s generation trajectories, we didn’t find a specific pattern. The model sometimes generates function words first, and sometimes generates content words first. Here’re two examples:
% Since we focus on text infilling here, our training objective (Eq. 6) does not favor a particular order, but encourages the model to realize a sentence equally well in all orders. This way, the model can flexibly start from any partial input text and complete it, regardless of the position of blanks. On the contrary, if the model always generates in a certain order, say first function words and then content words, it would have difficulty filling in the function words given an input consisting of content words. Depending on the application, we can also train the model to generate in specific orders by placing higher weights on the corresponding trajectories. We’ll add these discussions and more examples to the paper.

The action of BLM consists of selecting a blank and replacing it with a word and possibly adjoining blanks.
We train BLM by optimizing a lower bound on the marginal data likelihood that sums over all possible generation trajectories. 
In this way, we encourage the model to realize a sentence equally well in all orders, which is suitable for filling arbitrary blanks.
Appendix~\ref{sec:example_trajectories} shows examples generated by BLM along with their trajectories.
Depending on the application, we could also train the model to generate in specific orders by placing higher weights on the corresponding trajectories.

%In future work, we would like to develop our model for information fusion and assisting human writing.
BLM has plenty of future applications, including template filling, information fusion, assisting human writing, etc.
Moreover, we can extend our formulation to a conditional generative model.
Such models can be used in machine translation to support editing and refining translation, as well as in dialogue systems to compose a complete sentence with given elements.
While we proposed BLM for language generation, it would also be interesting to compare the representations learned by BLM with those produced by other pre-training methods.

%Future work may explore sequence modeling tasks beyond text rewriting that also benefit from flexible generation order. An example is music modeling: harmonic constraints naturally impose a canvas that composers fill in with the melody.

\section*{Acknowledgments}
We thank all reviewers and the MIT NLP group for their thoughtful feedback.

\bibliographystyle{acl_natbib}
\bibliography{emnlp2020}

\newpage
\appendix

\twocolumn[
\begin{center}
{\Large \bf
%Supplementary Material for \papertitle
Appendix
}
\vspace{40pt}
\end{center}
]

\section{Implementation Details for Text Infilling Baselines}

\subsection{Insertion Transformer}

We implement the Insertion Transformer in our own framework, using the same Transformer encoder module as for BLM and replacing the prediction layers by Insertion Transformer's mechanism. 
The canvas is also generated according to the training procedure of Insertion Transformer.

\subsection{Masked Language Model} 

We use the \texttt{RobertaForMaskedLM} architecture 
in the Transformers library
for MLM \cite{Wolf2019HuggingFacesTS, liu2019roberta}. 

At test time, the model is given an easier version of the text infilling task where blanks are expanded into sequences of $\langle$mask$\rangle$ tokens of the target length (or equivalently, the model uses an oracle to predict the length of the infilling).

We experiment with three decoding strategies: (1) one-shot: the model predicts all masks simultaneously (2) left-to-right: the model fills in the masks from left to right (3) confident-first: the model fills one mask at a time that has the highest score. We report results for the confident-first strategy which has the best performance.

\subsection{BERT+LM} 

We use the
\texttt{bert-base-uncased} model as served by the Transformers library \cite{Wolf2019HuggingFacesTS, devlin2018bert}.
The left-to-right language model is %either an LSTM or
a Transformer decoder to predict tokens in a blank.
Its input word embedding is concatenated with BERT's output in the blank position at each time step. 
%We report the results for the Transformer architecture.

\subsection{Seq2seq-full and Seq2seq-fill}

For both seq2seq baselines, we use Fairseq's  \texttt{transformer\_iwslt\_de\_en} architecture \citep{ott2019fairseq}.
To generate training data, we apply the blanking procedure to the input dataset and generate $k$ copies of each sentence with different masks. We experiment with $k \in \{1, 10, 100\}$ and report the best performance, obtained by $k = 10$.

\section{Monte-Carlo Estimate of Perplexity}\label{sec:mc}
For a sentence $x$ of length $n$, we estimate $p(x;\theta)$ in Eq. (\ref{eq:likelihood}) with $m$ samples:
$$X_m = \frac{n!}{m}\sum_{i=1}^m p(x, \sigma_i;\theta)$$ where $\sigma_i$'s are randomly sampled orders.

Note that $X_m$ is an unbiased estimate of $p(x;\theta)$:
$$\mathbb E[X_m] = p(x;\theta)$$

The estimated PPL is accordinly: $$Y_m = X_m^{-1/n}$$

Since $z^{-1/n}$ is a convex function of $z$,
$$\mathbb E[Y_m] = \mathbb E[X_m^{-1/n}] \ge \mathbb E[X_m]^{-1/n} = p(x;\theta)^{-1/n}$$
i.e., the expectation of the estimated PPL $\ge$ the actual PPL. As $m$ increases, the variance of $X_m$ decreases, and the inequality becomes tighter.

Hence, we will observe that as $m$ increases, the estimated PPL becomes smaller and converges to the real PPL. 
%We choose $m$ that we observe close to convergence on the dataset.

\clearpage

\onecolumn

\section{Generation Trajectory}\label{sec:example_trajectories}

\begin{figure}[ht]
\fontsize{10}{12}\selectfont
\centering
\begin{tabular}{l}
\toprule
% \underline{\qquad} \\
% \underline{\qquad} really \underline{\qquad} \\
% \underline{\qquad} really \underline{\qquad} ! \\
% \underline{\qquad} really \underline{\qquad} tea ! \\
% \underline{\qquad} really \underline{\qquad} tom \underline{\qquad} tea ! \\
% i \underline{\qquad} really \underline{\qquad} tom \underline{\qquad} tea ! \\
% i also really \underline{\qquad} tom \underline{\qquad} tea ! \\
% i also really \underline{\qquad} the tom \underline{\qquad} tea ! \\
% i also really \underline{\qquad} the tom yum tea ! \\
% i also really love the tom yum tea ! \\
% \midrule
\underline{\qquad} \\
\underline{\qquad} also \underline{\qquad} \\
the \underline{\qquad} also \underline{\qquad} \\
the \underline{\qquad} also \underline{\qquad} choice \underline{\qquad} \\
the salsa \underline{\qquad}also \underline{\qquad} choice \underline{\qquad} \\
the salsa was also \underline{\qquad} choice \underline{\qquad} \\
the salsa was also \underline{\qquad} only choice \underline{\qquad} \\
the salsa was also \underline{\qquad} only choice . \\
the salsa was also my only choice . \\
\midrule
\underline{\qquad} \\
\underline{\qquad} , \underline{\qquad} \\
\underline{\qquad} , \underline{\qquad} terrible \underline{\qquad} \\
\underline{\qquad} poor \underline{\qquad} , \underline{\qquad} terrible \underline{\qquad} \\
\underline{\qquad} poor \underline{\qquad} , \underline{\qquad} terrible , \underline{\qquad}\\
\underline{\qquad} poor \underline{\qquad} , \underline{\qquad} terrible , very \underline{\qquad} \\
\underline{\qquad} poor selection , \underline{\qquad} terrible , very \underline{\qquad} \\
very poor selection , \underline{\qquad} terrible , very \underline{\qquad} \\
very poor selection , service terrible , very \underline{\qquad} \\
very poor selection , service terrible , very \underline{\qquad} ! \\
very poor selection , service terrible , very slow ! \\
\midrule
\underline{\qquad} \\
\underline{\qquad} favorite \underline{\qquad} \\
my favorite \underline{\qquad} \\
my favorite \underline{\qquad} pittsburgh \underline{\qquad} \\
my favorite \underline{\qquad} pittsburgh . \\
my favorite restaurant \underline{\qquad} pittsburgh . \\
my favorite restaurant in pittsburgh . \\
\midrule
\underline{\qquad} \\
\underline{\qquad} the \underline{\qquad} \\
\underline{\qquad} is \underline{\qquad} the \underline{\qquad} \\
\underline{\qquad} is \underline{\qquad} the \underline{\qquad} . \\
\underline{\qquad} is \underline{\qquad} the \underline{\qquad} are \underline{\qquad} . \\
\underline{\qquad} food is \underline{\qquad} the \underline{\qquad} are \underline{\qquad} . \\
\underline{\qquad} food is \underline{\qquad} the \underline{\qquad} are \underline{\qquad} friendly . \\
\underline{\qquad} food is \underline{\qquad} and the \underline{\qquad} are \underline{\qquad} friendly . \\
\underline{\qquad} food is delicious and the \underline{\qquad} are \underline{\qquad} friendly . \\
\underline{\qquad} food is delicious and the \underline{\qquad} are very friendly . \\
\underline{\qquad} food is delicious and the owners are very friendly . \\
the food is delicious and the owners are very friendly . \\
\toprule
\end{tabular}
\caption{Examples of BLM generation trajectory on the Yelp review dataset.}
\label{fig:examples_lerp_yahoo}
\end{figure}

\end{document}